\documentclass[conference]{IEEEtran}
\IEEEoverridecommandlockouts
\usepackage{cite}
\usepackage{amsmath,amssymb,amsfonts}
\usepackage{algorithmic}
\usepackage{graphicx}
\usepackage{textcomp}
\usepackage{xcolor}
\def\BibTeX{{\rm B\kern-.05em{\sc i\kern-.025em b}\kern-.08em
    T\kern-.1667em\lower.7ex\hbox{E}\kern-.125emX}}

\usepackage{float}
\usepackage{float}
\usepackage{graphicx} 
\usepackage{subcaption}
\usepackage{amsmath}
\usepackage{dblfloatfix}

\vspace{-5pt} 
\usepackage{arydshln}
\date{January 2025}
\usepackage{booktabs} 
\begin{document}

\title{Assessing Human Cooperation for Enhancing Social Robot Navigation
}





\author{\IEEEauthorblockN{ Hariharan Arunachalam}
\IEEEauthorblockA{\textit{LAAS-CNRS} \\
\textit{Universite de Toulouse}\\
Toulouse, France \\
harunachal@laas.fr}
\and
\IEEEauthorblockN{Phani Teja Singamaneni}
\IEEEauthorblockA{\textit{LAAS-CNRS} \\
\textit{Universite de Toulouse}\\
Toulouse, France \\
ptsingaman@laas.fr}
\and
\IEEEauthorblockN{Rachid Alami}
\IEEEauthorblockA{\textit{LAAS-CNRS} \\
\textit{Universite de Toulouse}\\
Toulouse, France \\
alami@laas.fr}
}


 \maketitle
\begin{abstract}
    Socially aware robot navigation is a planning paradigm where the robot navigates in human environments and tries to adhere to social constraints while interacting with the humans in the scene. These navigation strategies were further improved using human prediction models, where the robot takes the potential future trajectory of humans while computing its own. Though these strategies significantly improve the robot's behavior, it faces difficulties from time to time when the human behaves in an unexpected manner. This happens as the robot fails to understand human intentions and cooperativeness, and the human does not have a clear idea of what the robot is planning to do. In this paper, we aim to address this gap through effective communication at an appropriate time based on a geometric analysis of the context and human cooperativeness in head-on crossing scenarios. We provide an assessment methodology and propose some evaluation metrics that could distinguish a cooperative human from a non-cooperative one. Further, we also show how geometric reasoning can be used to generate appropriate verbal responses or robot actions. 
    


\end{abstract}

\section{Introduction}
\label{sec:introduction}
Explainable robots are studied widely in literature \cite{setchi2020explainable}, and recent surveys \cite{mavrogiannis2023core, francis2025principles, hoffman2023inferring, singamaneni_ijrr_24} indicate that methodologies for inferring human intention are crucial for higher fluency in human-robot interaction. 

Recent work in \cite{fujioka2024need} develops a human behavior model in corridor cases. The authors identify that humans in general have different behaviors based on their understanding of the robot's intention and start to contribute to collision avoidance when the robot's intent is clear. In order to detect a human's understanding, the authors check a single point in the human path and select a robot behavior as a response.
Moreover, the scenario tested by the authors is specific to a narrow condition and doesn't account for the robot's reaction to a wider corridor. Although the authors conducted tests on nonverbal communication strategies based on the human's behavior inferred by the robot, an earlier work in \cite{nikolaidis2018planning} shows that verbal cues are more effective for collaborative tasks compared to only non-verbal actions.
 
This work tries to assess human behavior in head-on crossing scenarios based on motion, and proposes a decision strategy that takes a robot's plan and generates pertinent explanations when needed. It also introduces the calculation of useful attributes from a socially aware navigation planner\cite{singamaneni2021human}, which indicates the robot's intention. 
The proposed framework assesses the situation (physical space) and human contribution in a scenario based on some (new) metrics and generates multiple reasoning predicates. These predicates are then used to generate verbal cues that convey the robot's intention. Finally, experiments are conducted in four different head-on crossing scenarios to show the effectiveness of the proposed methodology in assessing human contribution. 


\section{Methodology}
\label{sec:framework}
\begin{figure}[H]
    \centering
    \includegraphics[width=\linewidth]{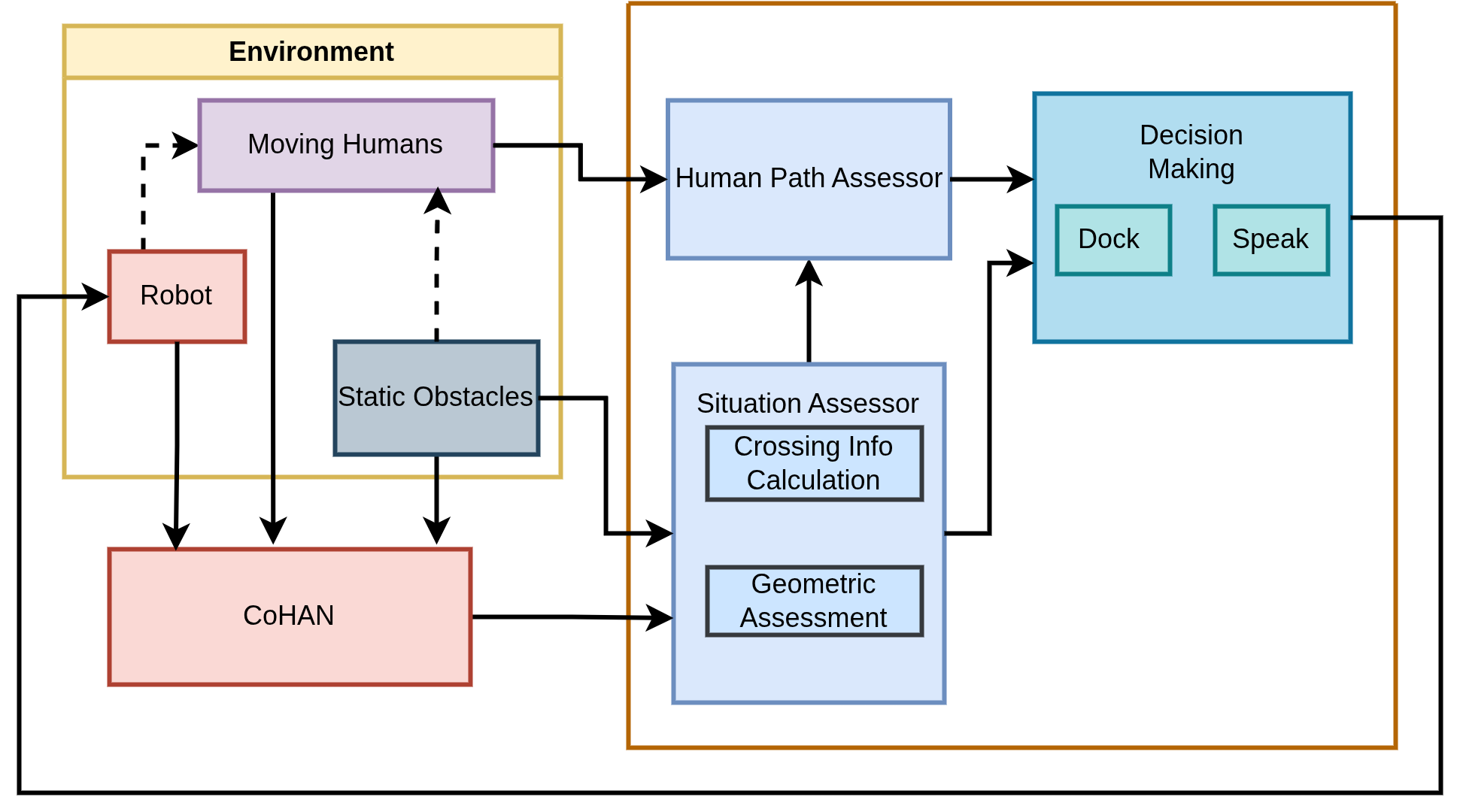}
    \caption{Proposed framework showing different modules of the assessment block. 
    }
    \label{fig:ecohan_framework}
\end{figure}
Fig. \ref{fig:ecohan_framework} shows the proposed framework, which consists of multiple components that work in synergy to provide explainable speech commands for the robot conditioned on the scenario. The main components of this framework are explained below.

\subsection{Use of CoHAN}
CoHAN is a socially aware navigation planner that plans trajectories for both a human and a robot, given their current states and goal states. More information on the functioning of CoHAN can be found in \cite{singamaneni2021human, singamaneni2022watch}. This work enhances CoHAN with the capability to inform the robot's intent verbally to the human. 

CoHAN outputs dual time-elastic bands \cite{rosmann2015timed, Khambhaita_rr_2020} (trajectories), one for the robot(\(\textbf{R}(t)\)) and one anticipated from the human(\(\textbf{H}(t)\)). These two trajectories are used to calculate different estimated attributes of an anticipated interaction. In this work, CoHAN is tuned to anticipate minimal contribution from the human, which pushes the robot to contribute more in a given scenario. These bands are optimized from an initial shortest path to the goal. 

\subsection{Situation Assessor}
The \textit{Situation Assessor} in the framework uses the known map of the environment and the two bands optimized by CoHAN to calculate the information for geometric reasoning. This module uses the crossing and geometric information to assess the navigation situation and produces three predicates that are later used for decision making: 
\begin{itemize}
    \item \(HumanNeedsToContribute\)
    \item \(HumanIsConstrained\)
    \item \(RobotIsConstrained\)
\end{itemize}


\subsubsection{Crossing Info Calculation}
Given two trajectories optimized by CoHAN \( \{\mathbf{R}(t_i)\}_{i=1}^N \) and \( \{\mathbf{H}(t_i)\}_{i=1}^N \), the time index \( i^* \) at which the agents are closest is given by: 

\[
i^* = \arg\min_{1 \leq i \leq N} \| \mathbf{R}(t_i) - \mathbf{H}(t_i) \|
\]

The corresponding points (Crossing Points) and time (time to cross) are:

\[
CP_h = \mathbf{H}(t_{i^*}) 
\]
\[
CP_r  = \mathbf{R}(t_{i^*})
\]
\[
t_{CROSS} = t_{i^*}
\]



Based on \(CP_h\) and \(CP_r\), the direction of crossing (\(Dir\)) is calculated. 
\subsubsection{Geometric Assessment}
\begin{figure}[ht]
    \centering
    \begin{subfigure}[b]{0.5\linewidth}
        \includegraphics[width=\linewidth]{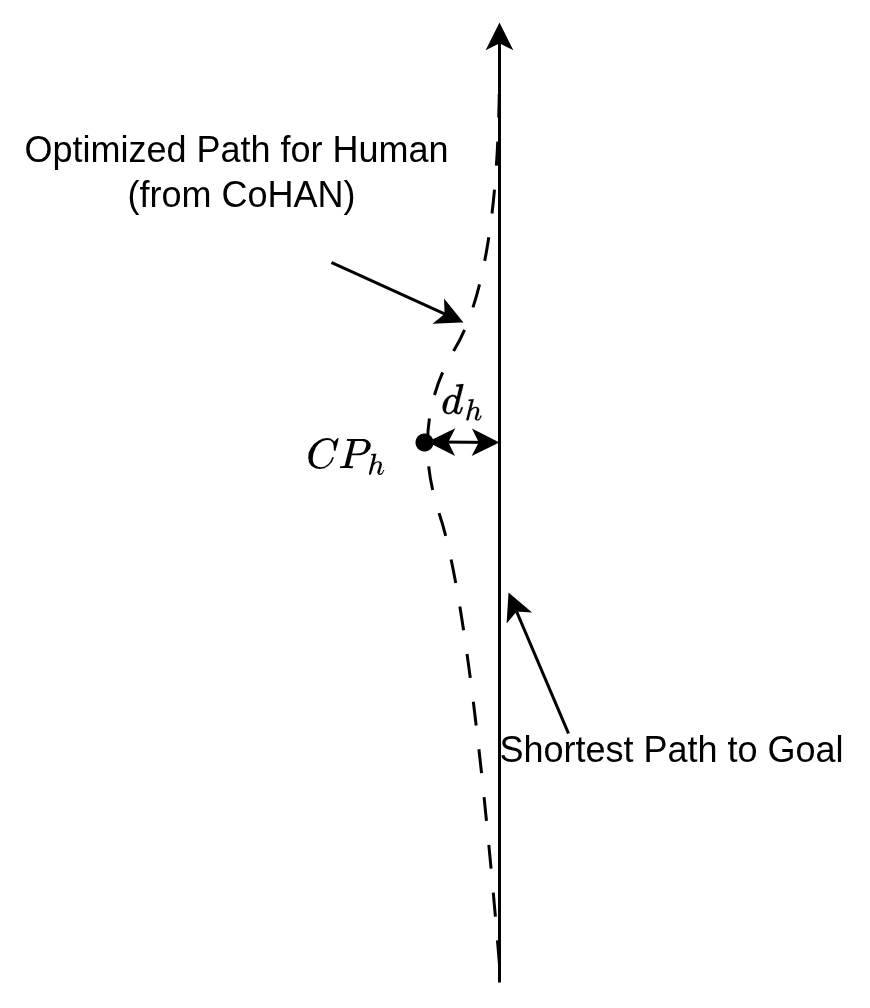}
        \caption{Path Comparison}
        \label{fig:tightband_comparison}
    \end{subfigure}
    \hfill
    \begin{subfigure}[b]{0.3\linewidth}
        \includegraphics[width=\linewidth]{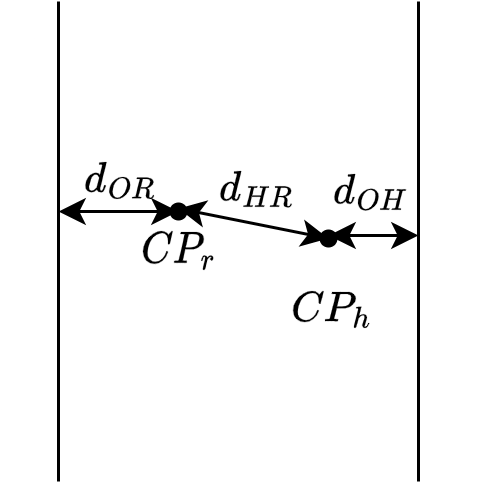}
        \caption{Distances at Crossing Point}
        \label{fig:distances_at_Cp}
    \end{subfigure}
    \caption{Situation Assessment}
    \label{fig:main}
\end{figure}

Here, the system first assesses if the human is expected to contribute at the crossing point. As mentioned earlier, the planner gives the least contributing trajectory for the human, which is intended to have the least deviation from the shortest path to the goal for the human. As shown in Fig.\ref{fig:tightband_comparison}, at the Crossing Point \(CP_h\), the anticipated human trajectory has a deviation of \(d_{h}\) from the shortest path to the goal. Therefore, the human needs to contribute (or cooperate) in this situation for the robot's navigation to be successful.
\[
HumanNeedsToContribute =  d_h > \tau_h
\]
\textcolor{black}{\(\tau_h\) in  the above equation is a threshold set on \(d_h\) which helps decide is the human needs to contribute.}

\noindent Apart from these, it evaluates the distances shown in Fig. \ref{fig:distances_at_Cp} to check the following: 
\begin{enumerate}
    \item \textit{The human has enough room to contribute}: Based on the distance to the closest obstacle \(d_{OH}\) (away from the robot) to the human at \(CP_h\).
\[
HumanIsConstrained = d_{OH} < \tau_{OH}
  \]  

\textcolor{black}{\(\tau_{OH}\) in  the above equation is a threshold set on \(d_{OH}\) which indicates if the human will be constrained in space to move or not.}
    
    \item \textit{The robot is constrained to move}: Since the robot is expected to contribute maximally, this is based on the distance to the closest obstacle \(d_{OR}\)
 (away from the human) at \(CP_r\) and the distance between human and robot at that point \(d_{HR}\).

\end{enumerate}

 \[
 RobotIsConstrained = (d_{HR} < \tau_{HR} \ \& \ \ d_{OR} < \tau_{OR}) 
\]

\textcolor{black}{Both \(\tau_{HR}\) and \(\tau_{OR}\) are distance thresholds on \(d_{HR}\) and \(d_{OR}\) for deciding if the robot will be able to maintain sufficient distance from both human and the closest obstacle respectively.}
\subsection{Human Path Assessor}

In a navigation scenario, any deviation away from the shortest path of the human helps us understand the cooperative nature of the human, where deviations which leads to increase in distance between the robot and the human is considered to be facilitating the navigation task and deviations which leads to reduction in the distance as non-contributing. Hence, this module assesses the human contribution and produces the \(IsContributing\) predicate, which is used in the decision-making. The following formulation is proposed to estimate the contributing nature of a given human.
\begin{figure}[H]
    \centering
    \includegraphics[width=0.7\linewidth]{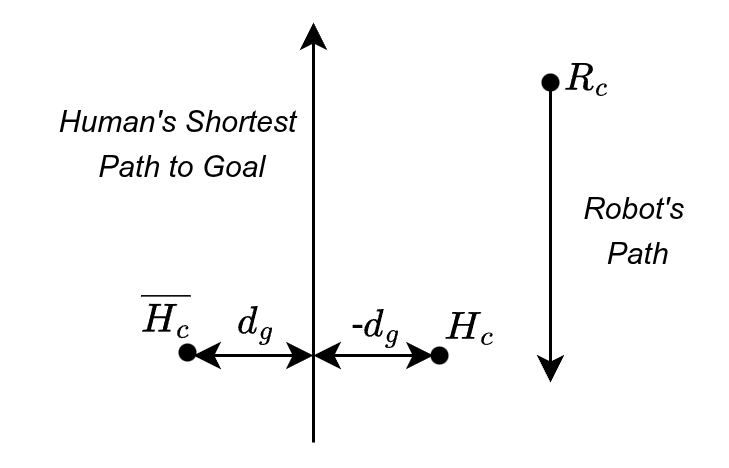}
    \caption{Signed Human Contribution}
    \label{fig:is_human_contributing}
\end{figure}

As shown in Fig.\ref{fig:is_human_contributing}, contribution is measured as the distance that the human moves with respect to the initial shortest path. The distance is signed based on the robot's location. The signed distance is used to create an array \(CA\) that records the human contribution over time.

\[
    CA =[ d_g^1  , .. , d_g^{N-1} , d_g^N]
\]

Since the question of \textbf{``when''} the contribution begins is highly subjective, a weighted average over these values \(CM\) is calculated, where the earlier points are given exponentially lesser weights as:

\[
CM = \frac{\sum_{i = 1}^{N} \gamma^{N-i} CA_i}{\sum_{i = 1} ^ {N} \gamma^{N-i}}  , \ 0 < \gamma < 1
\]
Finally, the decision about contribution is taken as follows:
\[
    IsContributing = \begin{cases}
        True \ , if \ CM > \tau 
        \\ 
        False \ , \ Otherwise
    \end{cases}
    \]
where $\gamma$ is the discount factor and $\tau$ is a chosen threshold on \(CM\).

\subsection{Decision Making}
\begin{figure*}[!t]
    \centering
    \includegraphics[width=0.9\linewidth]{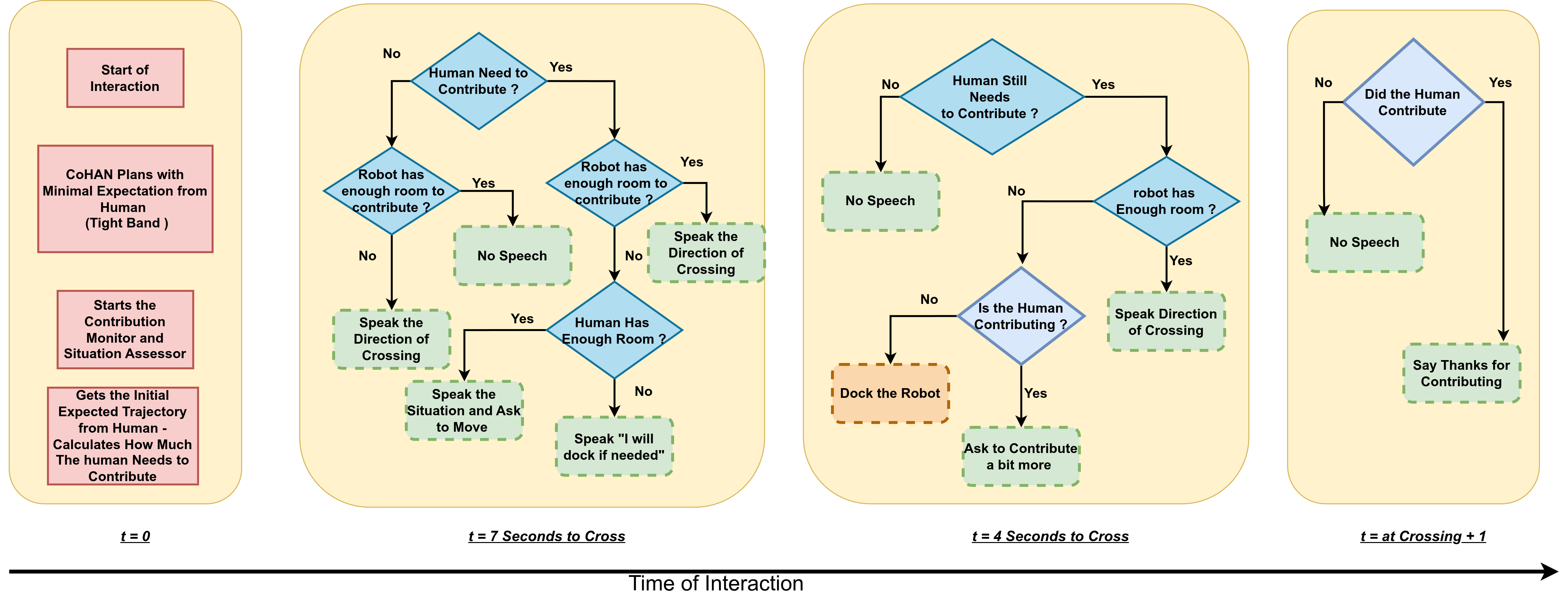}
    \caption{Pipeline}
    \label{fig:ecohan_pipeline}
\end{figure*}

The flowchart in Fig.\ref{fig:ecohan_pipeline} shows the coordination between different components in the framework and the decision making pipeline. The current pipeline has the assumption of knowing the human's goal prior to  execution. 
For the timeline of the interaction, the framework uses the estimated \(t_{CROSS}\).
The pipeline is described below :
\begin{enumerate}
    \item Once the scenario starts and \(t_{CROSS} \approx 7\), the decision making scheme starts. At this point, the system starts to record the human motion for assessment.
    \begin{itemize}
        \item if \(HumanNeedsToContribute\) is True, then the system checks if \(RobotIsConstrained\)
        \begin{itemize}
            \item if the robot has enough space, the robot informs the \(Dir\) (direction of crossing) to the human.
            \item but if the robot is constrained, then the system checks if \(HumanIsConstrained\)
            \begin{itemize}
                \item if the human has enough room, then the robot informs the situation and suggests a direction for the human to move.
                \item if the human is also constrained in space, it will indicate that the robot will dock if needed.
            \end{itemize}
        \end{itemize}
        \item If the human is not constrained, and 
        \begin{itemize}
            \item if the robot is constrained, the robot speaks the direction of crossing.
            \item in all other cases, the robot doesn't speak
        \end{itemize}
    \end{itemize}

    \item As the scenario progresses and at \(t_{CROSS} \approx  4\), the system does another round of assessment and decisions. The system checks if the human still needs to contribute: 
        \[
        StillNeedsToContribute = CM < d_h
        \]
    \begin{itemize}
        \item if the human still needs to contribute, and 
        \begin{itemize}
            \item if the robot still has enough space at the crossing, then the robot informs the intended crossing direction to the human again.
            \item else if the robot is constrained, it checks if the human has been contributing with the Human Path Assessor and resets the distance recorder.
            \begin{itemize}
                \item if the human \(IsContributing\), the robot asks the human to move a bit more.
                \item else the robot decides to dock to its nearest wall away from the human.
            \end{itemize}
        \end{itemize}
        \item if the human doesn't need to contribute anymore, the robot doesn't say anything.
    \end{itemize}
    \item Once the robot and the human cross each other, the system stops the Human Path Assessor again and checks if the human has facilitated the crossing. If the human has facilitated the crossing, the robot thanks the human for cooperation.
    
\end{enumerate}

\section{Experiments and Discussion}
\begin{figure}[H]
    \centering
    \begin{subfigure}[b]{0.49\linewidth}
         \includegraphics[width=\linewidth]{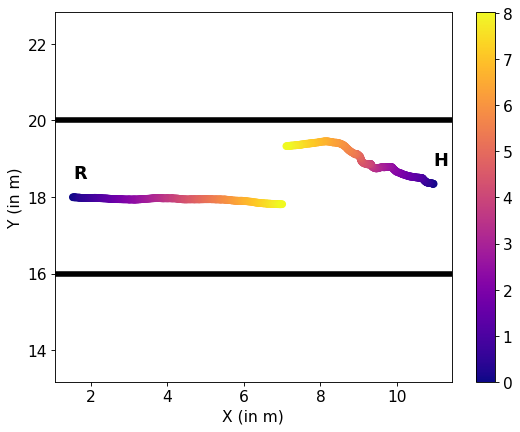}
        \caption{Facilitating}
        \label{fig:cooperative_open}
    \end{subfigure}
    \hfill
    \begin{subfigure}[b]{0.49\linewidth}
        \includegraphics[width=\linewidth]{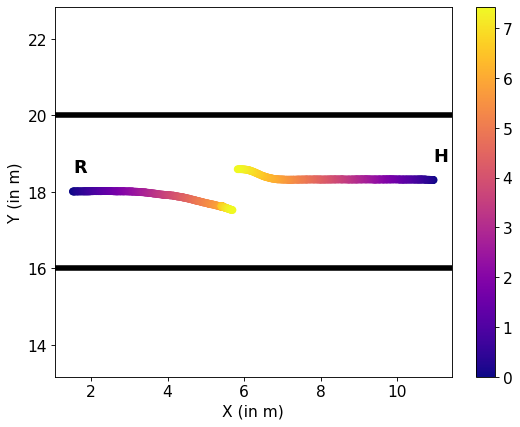}
        \caption{Minimally Contributing}
        \label{fig:minimal_open}
    \end{subfigure}
    \caption{Open Corridor}
    \label{fig:open_space}
\end{figure}

 \begin{figure}[H]
    \centering
    \begin{subfigure}[b]{0.49\linewidth}
    \includegraphics[width=\linewidth]{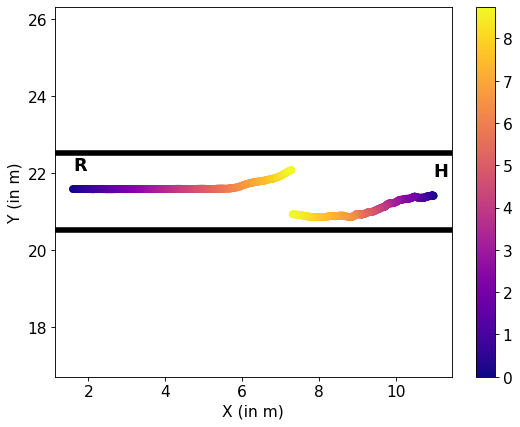}
        \caption{Facilitating}
        \label{fig:cooperative_tight}
    \end{subfigure}
    \hfill
    \begin{subfigure}[b]{0.49\linewidth}
        \includegraphics[width=\linewidth]{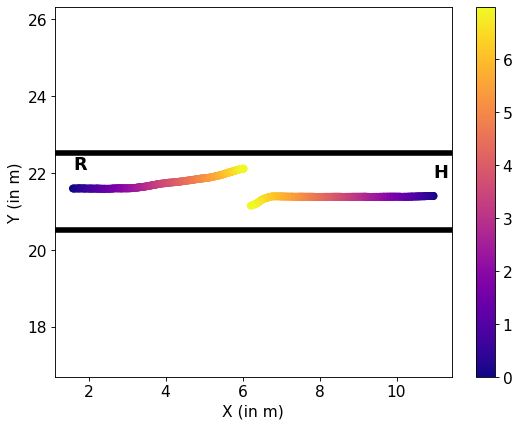}
        \caption{Minimally Contributing}
        \label{fig:minimal_tight}
    \end{subfigure}
    \caption{Narrow Corridor}
    \label{fig:tight_space}
\end{figure}

\label{sec:experiments}
 The proposed framework is run in two different environments with two different humans in each environment totaling 4 scenarios: 
 \begin{itemize}
     \item Open Corridor with Facilitating Human (Fig.~\ref{fig:cooperative_open})
     \item Open Corridor with Minimally Contributing Human (Fig.~\ref{fig:minimal_open})
     \item Narrow Corridor with Facilitating Human (Fig.~\ref{fig:cooperative_tight})
     \item Narrow Corridor with Minimally Contributing Human (Fig.~\ref{fig:minimal_tight})
 \end{itemize}

In all the cases, the human and the robot have goals behind each other's starting positions. The `Open Corridor' has a separation of more than 3 meters, and a `Narrow Corridor' has less than 3 meters of separation. The Minimally Contributing Human avoids the robot as an obstacle, but only at the last moments before crossing. The Minimally Contributing Human is simulated with TEB \cite{rosmann2015timed}, and a Facilitating Human runs another instance of CoHAN, which makes the human to move away from (or facilitate) the robot much sooner. 

In order to provide enough latitude for the human in the scenario, \( \tau_{OH}\) is set to 1 meter, and in order to avoid collision with the obstacle, \(\tau_{OR}\) is set to 0.3 meters. Further, to maintain a social distance at the crossing point, \(\tau_{HR}\) is set to 1.2 meters.

\subsection{Detecting Cooperative Behavior}

The trajectory plots in Fig.~\ref{fig:open_space} and Fig.~\ref{fig:tight_space} show the actual trajectory followed by the robot and the human till they cross each other. By observing these figures, it is clear that the human has facilitated the crossing in two cases, while in the other two, only collision avoidance was performed.  The Human Path Assessor described in Sec. \ref{sec:framework} uses these points to check if the human is contributing.

\begin{figure}[h]
    \centering
    \includegraphics[width=0.8\linewidth]{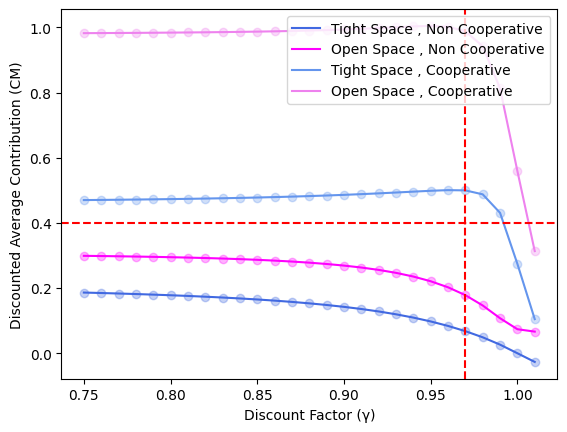}
    \caption{\( \gamma\) vs \(CM\)}
    \label{fig:gamma_vs_tau}
\end{figure}

The Path Assessor has two design parameters for deciding if the human is contributing, \(\gamma\) and \(\tau\), \(\gamma\) modifies the importance of contribution temporally, that is \(\gamma < 1\)  ensures that the latest contributions are weighted more in the average and when \(\gamma > 1 \)  makes later contributions weigh less in the average. The plot in Fig.\ref{fig:gamma_vs_tau} shows the average with different \(\gamma\) values.

With the current examples shown in the plots, \(\gamma\) value of 0.98 differentiates the Facilitating human from the Minimally Contributing one,  and a \(\tau\) of 0.4 is selected on the discounted average to decide if the human is contributing.

\begin{table}[h]
\centering
\begin{tabular}{|c |c |c |} 
 \hline
 Scenario & Minimally Contributing & Facilitating \\ 
 \hline
 Open & 0.15 & 0.94  \\ 
 \hline
 Narrow & 0.05 & 0.49 \\
 \hline
\end{tabular}
\caption{\(CM\) values for different scenarios}
\label{tab:cm_values}
\end{table}

\begin{figure}[!h]
    \centering
    \begin{subfigure}[b]{0.49\linewidth}
        \includegraphics[width=\linewidth]{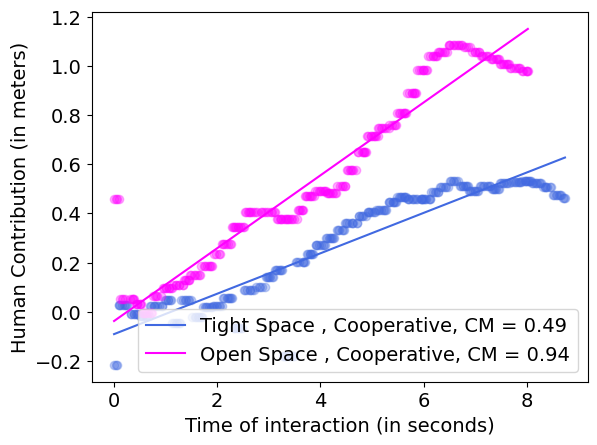}
        \caption{Facilitating}
        \label{fig:sub1}
    \end{subfigure}
    \hfill
    \begin{subfigure}[b]{0.49\linewidth}
        \includegraphics[width=\linewidth]{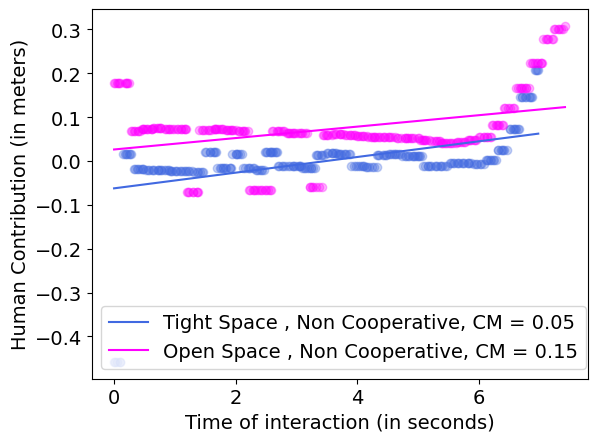}
        \caption{Minimally Contributing}
        \label{fig:sub2}
    \end{subfigure}
    \caption{Contributing Distances (\(CA\))}
    \label{fig:contribution_array}
\end{figure}

The plots in Fig.~\ref{fig:contribution_array} show values of the array, \(CA\), used by Path Assessor, and Table~\ref{tab:cm_values} shows the \textit{CM} values calculated with the current design parameters. From the table, it can be observed that the value \(CM\) reduces as human contribution reduces in the scenario.


\section{Conclusion}
\label{sec:conclusion}
This work proposed an initial framework for generating verbal cues for corridor navigation scenarios with different human behaviors, and proposes a assessment technique to classify a human's behavior based on human path. The framework sometimes decides to dock the robot, and more such non-verbal cues can be mapped to the observed human behavior. Moreover, eventhough the assessment technique is preliminary, it shows a differentiable feature that can be made a learnable parameter in the future work. Further user studies will be conducted to confirm the pertinence of the proposed method.

\bibliographystyle{ieeetr}
\bibliography{reference}

\end{document}